# Comparison of Guidance Modes for the AUV "Slocum Glider" in Time-Varying Ocean Flows

Mike Eichhorn
Institute for Automation and Systems Engineering
Ilmenau University of Technology
98684 Ilmenau, Germany
Email: mike.eichhorn@tu-ilmenau.de

Hans Christian Woithe, Ulrich Kremer
Department of Computer Science
Rutgers University
Piscataway, New Jersey 08854
Email: {hcwoithe,uli}@cs.rutgers.edu

*Abstract* — **This paper presents possibilities for the reliable guidance of an AUV "Slocum Glider" in time-varying ocean flows. The presented guidance modes consider the restricted information during a real mission about the actual position and ocean current conditions as well as the available control modes of a glider. A faster-than-real-time, full software stack simulator for the Slocum glider will be described in order to test the developed guidance modes under real mission conditions.**

## I. Introduction

To guide an AUV from a starting location to a destination location a lot of information is required, such as the area of operation (e.g., shipping lanes, ocean currents, areas to avoid), in addition to information about the vehicle's behavior and its status. This information is not constant over space and time. Planning a future glider mission by a human pilot requires extensive experience to interpret all the information. An automated path planning system could free operators from the tedious task of waypoint selection and would allow them to focus on scientific and mission critical aspects of managing groups of AUVs.

There exist a variety of solutions for path planning in a time-varying ocean flow for AUVs. It is a challenge to translate the discovered path into executable commands for the AUV. In [1], the level set method for time-optimal path planning is used. The numerical solution provides an optimal heading time-series for the AUV. A modified A* algorithm named Constant-Time Surfacing A* (CTS-A*) generates a sequence of bearing angles to command a glider [2, 3]. These bearing angles take a drift correction into account in order to obtain the appropriate heading to reach the desired waypoint. Over the past years, we have been developing path planning algorithms based on graph methods to find a time-optimal path for the AUV "Slocum Glider" [4]. This research focused on the inclusion of practice-oriented considerations in planning, the acceleration of the algorithms and their usage to support glider pilots. In this paper we present possibilities to use a generated path under real preconditions on a glider.

The Slocum Electric Glider is an AUV developed and produced by Teledyne Webb Research [5]. It is a buoyancy driven AUV with an engine at the front of the vehicle which moves a piston to change the vehicle's water displacement and thereby its buoyancy. The pitch of the AUV can also be changed by moving an internal battery pack, thereby shifting the vehicle's center of gravity. Due to a set of wings mounted on both sides of the vehicle, the vehicle glides through the water following a saw-toothed flight profile, changing its buoyancy near the surface and at particular water depths. The Slocum glider belongs to a class of buoyancy driven AUVs which includes vehicles such as Bluefin Robotics' Spray Glider [6] and iRobot's Seaglider [7]. A Slocum glider with a double payload bay is shown in Fig. 1.

Using a rudder and the global positioning system (GPS), the glider is able to navigate and collect data samples using onboard sensors. Satellite and radio communications are used at the surface to transfer scientific and vehicle data, and if necessary, to alter the AUV's mission [8]. The Slocum glider has an approximate speed of 0.35 m/s, which is significantly slower than a typical propeller driven vehicle. However, the glider has the advantage that its buoyancy engine is required only during inflection points, making it a much more energy efficient vehicle. This allows prolonged flights typically lasting weeks or even months [8] depending on sensor payload and sensor usage. In contrast, typical mission durations of propeller driven vehicles are in terms of hours or a few days at best.

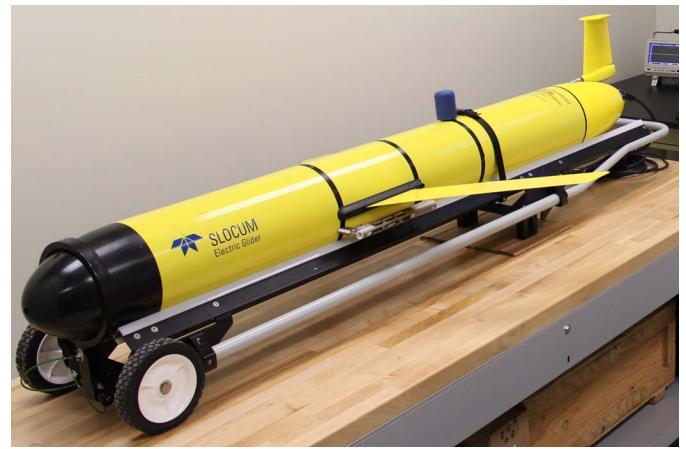

Fig. 1. A Slocum Glider with a double payload bay and a top-mounted acoustic modem in its deployment cart.

## II. Glider Guidance Modes

Section III presents a faster-than-real-time, full software stack simulator for the Slocum glider to support the glider pilot in order to verify his planned mission. The simulator includes vehicular, environmental, (e.g., CODAR, seafloor), ocean current models based on forecast information and energy models to determine the behavior of a glider while flying from one waypoint to the next [9]. This mode of operation is called **half automatic planning mode**, where the pilot designs a mission plan and verifies it in the simulator. Detailed investigations about this mode are presented in [9].

Another mechanism to generate a mission plan for a glider is by using a path planning algorithm. The goal of the path planning algorithm used in this paper is the finding of a time-optimal path from a start position to a goal position by evading all static and dynamic obstacles in the area of operation, while considering the dynamic behavior of the vehicle and the time-varying ocean current. This path planning algorithm, named A* Time Variant Environment (A*TVE) algorithm [10, 11, 12], is based on a modified A* algorithm. The path algorithm uses a geometric graph for the description of the area of operation with all of its characteristics. The defined points (vertices) within the operational area are those passable by the vehicle. The passable connections between these points are recorded as edges in the graph. Every edge has a rating (cost, weight) which is the time required for traversing the connection. In the case of an ocean current, the mesh structure of the geometric graph will be a determining factor associated with its special change in gradient. The path planning algorithm also uses a simple simulator to calculate the costs of the graph's edges. This mode is called **automatic planning mode**.

Such a path planning algorithm generates a waypoint list under the presumption that the vehicle has an accurate navigation system to determine its position underwater such that it can follow the commanded waypoints along the path elements.

Gliders use a dead reckoning (DR) algorithm to estimate their position underwater. This DR algorithm uses the data of the onboard pressure and attitude sensor to determine the current depth, pitch, roll and heading of the vehicle. Based on these data, the vehicle speed through water $v_{veh\_bf}$ and the earth fixed velocity $\mathbf{v}_{veh\_ef}$ are calculated. The position in local mission coordinates (LMC) can be obtained by integration of the velocity $\mathbf{v}_{veh\_ef}$. The inclusion of ocean current information is also possible, assuming that this information is reliable. This information is provided by the current correction (CC) system. The CC system generates a depth averaged ocean current using the last GPS update. Thus, the CC system is unsuitable for our test scenarios because we consider a time-varying ocean flow and different current conditions in several depths. In addition, it is a requirement in our test cases that the glider does not execute a GPS update during the whole mission.

Hence the CC is disabled in our tests and the DR calculation assumes that the ocean current velocity is zero. To ensure a correct operation of the heading algorithm in the glider software, the path planning waypoint list has to be modified. TABLE I includes the pseudo-code to transform a generated waypoint list in a DR waypoint list.

The first step includes the calculation of the travel time $t_{travel}$ to arrive at the next waypoint $WP[i]$ in function CALC-TRAVELTIME which is presented in detail in [10]. This function can also supply an optimal "diveto" depth and "climbto" depth for the path element. An optimal adaption of the glider dive profile is useful in regions with an adverse surface or seabed current. This additional extension of the CALC-TRAVELTIME function is described in detail in [11] and [13]. The function DETECT-HEADING (TABLE I, gray highlighted) is used to determine a command heading $\varphi$ which guides the glider to the next waypoint. Possibilities to determine this heading are presented in detail in section V. The next DR waypoint $WP\_DR[i]$ is determined using this heading $\varphi$, the calculated travel time $t_{travel}$ and the glider speed $v_{veh\_bf}$. In a final step, the glider track will be simulated in function CALC-DESTINATION using heading $\varphi$, having started on the current position $\mathbf{x}_{start}$ at the time $t_{start}$. The position $\mathbf{x}_{end}$ after a simulation period $t_{travel}$ is used as the start position in the next iteration.

At the end of this process, a DR waypoint list is generated which will be used in the glider software to guide the vehicle in a time-varying ocean current field along the path generated from a path planning system.

A glider may also be guided by defining a list of times for heading changes. The times correspond with the start time $t_{start}$ and the heading commands that are the results of the DETECT-HEADING function in TABLE I. Hence, the glider works off the heading list similar to a sequential control. The dead reckoning algorithm on the glider will not be used in this mode.

TABLE I
PSEUDO-CODE OF THE ALGORITHM TO CREATE A DEAD RECKONING WAYPOINT LIST

---

CREATE-DEAD-RECKONING-WAYPOINT-LIST($WP$, $t_0$)

*defined parameters:* $v_{veh\_bf}$, $z_{climb-to}$, $z_{dive-to}$

$\mathbf{x}_{start} = WP[1]$

$t_{start} = t_0$

$WP\_DR[1] = \mathbf{x}_{start}$

**for** ($i = 2$) **to** ($i$ = length($WP$))

  $\mathbf{x}_{end\_path} = WP[i]$

  $t_{travel}, z_{dive-to}, z_{climb-to}$ = CALC-TRAVELTIME($\mathbf{x}_{start}, \mathbf{x}_{end\_path}, t_{start}$)

  $t_{end} = t_{start} + t_{travel}$

  $\varphi$ = DETECT-HEADING($\mathbf{x}_{start}, \mathbf{x}_{end\_path}, t_{start}, t_{end}, z_{climb-to}, z_{dive-to}$)

  $WP\_DR[i] = WP\_DR[i-1] + \begin{bmatrix} \cos(\varphi) \\ \sin(\varphi) \end{bmatrix} t_{travel} v_{veh\_bf}$

  $\mathbf{x}_{end}$ = CALC-DESTINATION($\mathbf{x}_{start}, t_{start}, \varphi, t_{travel}, z_{climb-to}, z_{dive-to}$)

  $\mathbf{x}_{start} = \mathbf{x}_{end}$

  $t_{start} = t_{end}$

**return** $WP\_DR$

## III. GLIDER SIMULATOR

The waypoint lists generated using the guidance modes are tested with a full software stack glider simulator [9]. This simulator is a port of the vehicle's control software and is capable of running on commodity hardware. Like the actual vehicle, the simulator accounts for the complex interactions of sensors, motors and software components. Furthermore, it has been retrofitted to include vehicular, environmental, and energy models to portray a glider executing a mission in a virtual environment.

Traditionally, a mission is simulated using either -a physical glider in simulation mode, a "Shoebox" simulator, or a "Pocket" simulator. The "Shoebox" and "Pocket" simulators use a subset of the glider's electronics to execute the vehicle's control software. The former is a more complete development environment and uses more components, while the latter uses only the bare minimum. Common among all of the manufacturer provided simulators are that they run in real-time, which can make testing cumbersome, if not impossible. The new simulator, however, does not require any glider hardware and has also been extended to optionally run faster-than-real-time. This makes simulations that require glider control behavior more feasible.

To accomplish simulations, the vehicle's system makes use of the "simdrvr" device driver that is part of the standard control software. It is used whether the simulator is physical, or purely software, as in the software port. The driver can run multiple times in a given glider cycle, which lasts approximately four seconds. The driver will update vehicle and environmental information such as the glider's pitch angle and water current information. A separate component of "simdrvr" also runs periodically as an interrupt service routine (ISR) if the glider's motors are being simulated. This ISR generates simulated motor movement and voltage levels for other components of the software to use.

The "simdrvr" device driver is separate from the rest of the control software. The bulk of the system is unaware of whether the sensor readings and actuators are physically or virtually based. Thus, the glider attempts to work backwards, by dead reckoning and estimating its movement and position. At any given time, the "simdrvr" has knowledge of the actual position in the virtual environment, while the remaining part of the system only has knowledge of the dead reckoned position. Because of this discrepancy, it is vital for path planning systems to have knowledge of both the actual glider position and the position where the vehicle believes it is. Since the ported simulator is the glider's control software plus environmental information, it is ideal for the evaluation of the planning system.

TABLE II
GLIDER SIMULATOR PARAMETERS

| Acceleration | 30 mission hours in one minute |
|---|---|
| Programming Language | C |
| Compiler | gcc 4.6.3 |
| Operating System | Ubuntu Linux 12.04 |

## IV. SIMPLE SIMULATOR

This section describes a simple simulator to simulate glider missions using a DR waypoint list. This simulator includes the handling of the DR waypoint list to extract the command headings, which will be used in a glider track calculation. This calculation is the core of this simulator and is used also in section V to determine the heading command.

### A. Extract the control parameters

The determination of the destination waypoint requires a simulation of the glider track in a time-varying ocean current using a defined heading command $\varphi$ and a time period $t_{travel}$ where the glider holds the commanded heading. These parameters will be extracted from a DR waypoint list (see section II) for the individual waypoints $WP\_DR[i]$ according to the following equations:

$$\mathbf{v}_{dir} = WP\_DR[i] - WP\_DR[i-1] \qquad (1)$$

$$\varphi = \text{atan2}(\mathbf{v}_{dir}^{y}, \mathbf{v}_{dir}^{x}) \qquad (2)$$

$$t_{travel} = \|\mathbf{v}_{dir}\|/v_{veh\_bf} \qquad (3)$$

### B. Calculation of the destination in time-varying ocean flow

The simulation is based on a step size control for efficient calculation of numerical solutions of differential equations. Such an approach was also used in [10] for the calculation of the travel time and in [12] for the calculation of the optimal path direction. The step size control leads to a performance-enhancement of over 30% compared to a fixed step size in these applications. The idea is to simulate the glider track by starting on the start position $\mathbf{x}_{start}$. The simulation will be stopped if the simulated travel time $t_{start\_local}$ is larger than the defined travel time $t_{travel}$. The glider position can be calculated by a discrete integration of the glider velocity vector in earth fixed coordinates $\mathbf{v}_{veh\_ef}$ using the variable sample time $\Delta t$. This velocity vector $\mathbf{v}_{veh\_ef}$ will be calculated according to:

$$\mathbf{v}_{veh\_ef} = \mathbf{v}_{current} + \begin{bmatrix} \cos(\varphi) \\ \sin(\varphi) \end{bmatrix} v_{veh\_bf} \qquad (4)$$

The sample time will be adapted according to the resulting step size $h$. The calculation of the glider track includes the following steps:

1. Rough approximation of the velocity $\mathbf{v}_{rough\_veh\_ef}$ using only the current $\mathbf{v}_{current\_start}$ from the start point $\mathbf{x}_{start\_local}$ to the time $t_{start\_local}$.

2. Calculation of the end point $\mathbf{x}_{end\_local}$ with the calculated velocity $\mathbf{v}_{rough\_veh\_ef}$ and a defined time period $\Delta t$ which corresponds with $ht_{ravel}$.

3. Determination of the ocean current $\mathbf{v}_{current\_end}$ from the end point $\mathbf{x}_{end\_local}$ to the time $t_{start} + t_{start\_local} + \Delta t$.

4. Calculation of an average ocean current $\mathbf{v}_{current\_mean}$ during the time period $\Delta t$ by arithmetic mean of the two velocities $\mathbf{v}_{current\_start}$ and $\mathbf{v}_{current\_end}$.

5. Improved approximation of the velocity $\mathbf{v}_{improved\_veh\_ef}$ using the mean current $\mathbf{v}_{current\_mean}$.

This is followed by the calculation of the local error $error_{local}$ between the difference of the two velocities $\mathbf{v}_{rough\_veh\_ef}$ and $\mathbf{v}_{improved\_veh\_ef}$ and the determination of the new step size $h$ using the following equation for an optimal step size for a second order method [14]:

$$h = \max\left\{h_{min}, \min\left\{h_{max}, \tau h \sqrt{\frac{\varepsilon}{error_{local}}}\right\}\right\} \quad (5)$$

The parameter $\tau$ is a safety factor ($\tau \in (0, 1]$). Acceptance or rejection of this step will depend on the local error $error_{local}$ to a defined tolerance $\varepsilon$ and the calculated step size $h$ to the minimal step size $h_{min}$. TABLE III includes the details of the algorithm.

TABLE III
PSEUDO-CODE OF THE ALGORITHM TO CALCULATE THE DESTINATION

---

CALC-DESTINATION($\mathbf{x}_{start}$, $t_{start}$, $\varphi$, $t_{travel}$, $z_{climb\text{-}to}$, $z_{dive\text{-}to}$)
*defined parameters:* $v_{veh\_bf}$, $h$, $h_{min}$, $h_{max}$, $\varepsilon$, $\tau$

$t_{start\_local} = 0$
$\mathbf{x}^{2D}_{start\_local} = \mathbf{x}^{2D}_{start}$
$\mathbf{x}^{z}_{start\_local} = z_{climb\text{-}to}$
$\mathbf{v}_{veh\_bf} = \begin{bmatrix} \cos(\varphi) \\ \sin(\varphi) \end{bmatrix} v_{veh\_bf}$
$\Delta z = z_{dive-to} - z_{climb-to}$
$\mathbf{v}_{current\_start}$ = GET-CURRENT($\mathbf{x}_{start\_local}$, $t_{start}$)
**while** ($t_{start\_local} < t_{travel}$)
    $\mathbf{v}_{rough\_veh\_ef} = \mathbf{v}_{current\_start} + \mathbf{v}_{veh\_bf}$
    $\Delta t = h t_{travel}$
    $\mathbf{x}^{2D}_{end\_local} = \mathbf{x}^{2D}_{start\_local} + \Delta t \mathbf{v}_{rough\_veh\_ef}$
    $\mathbf{x}^{z}_{end\_local} = \mathbf{x}^{z}_{start\_local} + \Delta z (t_{start\_local} + \Delta t)/t_{travel}$
    $\mathbf{v}_{current\_end}$ = GET-CURRENT($\mathbf{x}_{end\_local}$, $t_{start} + t_{start\_local} + \Delta t$)
    $\mathbf{v}_{current\_mean} = 0.5(\mathbf{v}_{current\_start} + \mathbf{v}_{current\_end})$
    $\mathbf{v}_{improved\_veh\_ef} = \mathbf{v}_{current\_mean} + \mathbf{v}_{veh\_bf}$
    $error_{local} = \|\mathbf{v}_{rough\_veh\_ef} - \mathbf{v}_{improved\_veh\_ef}\|$
    $h = \max(h_{min}, \min(h_{max}, \tau h \sqrt{\varepsilon / error_{local}}))$
    **if** (($error_{local} < \varepsilon$) OR ($h = h_{min}$))
        $\mathbf{v}_{current\_start} = \mathbf{v}_{current\_end}$
        $\mathbf{x}^{2D}_{end\_local} = \mathbf{x}^{2D}_{start\_local} + \Delta t \mathbf{v}_{improved\_veh\_ef}$
        $\mathbf{x}^{2D}_{start\_local} = \mathbf{x}^{2D}_{end\_local}$
        $\mathbf{x}^{z}_{start\_local} = \mathbf{x}^{z}_{end\_local}$
        $t_{start\_local} = t_{start\_local} + \Delta t$
**return** $\mathbf{x}_{start\_local}$

---

The glider dives in this simulation from the "climbto" to the "diveto" depth. This allows the inclusion of the glider behavior in every passible depth in the calculations. In case of long travel times or a strong space and time varying ocean current the travel time $t_{travel}$ can be divided into several time intervals in which the algorithm described above runs repeatedly. This way the end position of the previous simulation is the start position of the next simulation. At this stage, we divide the travel time into time intervals of ten. This principle is also used in section V.B.

## V. HEADING DETERMINATION

The path planning algorithms used generate each path element on the assumption that the glider is able to follow them very accurately. If a position and time-varying ocean current appears the commanded heading $\varphi$ will be changed along the path element so that the glider can hold the track. Because the glider software requires only a single heading value to guide a glider to the next waypoint, this section describes three possible ways to detect such a single heading command $\varphi$ using a path element generated from a path planning algorithm. A path element is defined by a start point $\mathbf{x}_{start\_path}$ and a start time $t_{start}$, where the glider begins to follow the defined path element, as well as an end point $\mathbf{x}_{end\_path}$ with the corresponding end time $t_{end}$. During the drive, the glider dives a saw-tooth profile characterized by a "climbto" depth $z_{climb\text{-}to}$ and a "diveto" depth $z_{dive\text{-}to}$. These are the inputs of the general algorithm to detect the heading command $\varphi$.

$\varphi$ = DETECT-HEADING($\mathbf{x}_{start\_path}$, $\mathbf{x}_{end\_path}$, $t_{start}$, $t_{end}$, $z_{climb\text{-}to}$, $z_{dive\text{-}to}$)

which is used in TABLE I to generate the dead reckoning waypoint list.

### A. Heading Calulation Method

The first method is based on the calculation of the speed $v_{path\_ef}$ the vehicle travels with on the path in relation to a fixed world coordinate system. This calculation is used in the path planning algorithms to determine the travel time $t^i_{path}$ for the path elements. The relations of the ocean current vector $\mathbf{v}_{current}$ and the path/course vector $\mathbf{v}_{path}$ to the glider speed vector $\mathbf{v}_{veh\_bf}$ allow the calculation of the commanded glider heading (A detailed description is presented in section 4.1 in [9]).

TABLE IV includes the steps of this method. An average ocean current $\mathbf{v}_{current}$ is calculated in (6), accounting for the different ocean current conditions in each path element. The speed of the glider to follow the path in relation to an earth fixed world coordinate system $v_{path\_ef}$ can be determined by an intersection point between a line and a circle. The direction of this line is defined by a unit vector $\mathbf{v}^0_{path}$ of a point subtraction of the two-dimensional end point $\mathbf{x}^{2D}_{end}$ and start point $\mathbf{x}^{2D}_{start}$ in (7). The circle center corresponds to the current vector $\mathbf{v}_{current}$ and the radius corresponds to the cruising speed of the glider $v_{veh\_bf}$ (see Fig. 2).

Equation (8) and (9) include the calculations for $v_{path\_ef}$, which is the magnitude of the course vector $\mathbf{v}_{path}$. If the discriminant *disc* in (8) becomes negative, this means that the vehicle can no longer be held in that path. In this case, the discriminant *disc* will be defined as zero and the resulting glider heading is perpendicular to the path so that the drift to the desired path is minimal.

The course vector $\mathbf{v}_{path} = v_{path\_ef} \mathbf{v}^0_{path}$ will be used together with the ocean current vector $\mathbf{v}_{current}$ to calculate the glider speed vector $\mathbf{v}_{veh\_bf}$ in (10). The magnitude of this vector corresponds with the cruising speed of the glider $v_{veh\_bf}$, its direction characterizes the sought glider heading $\varphi$.

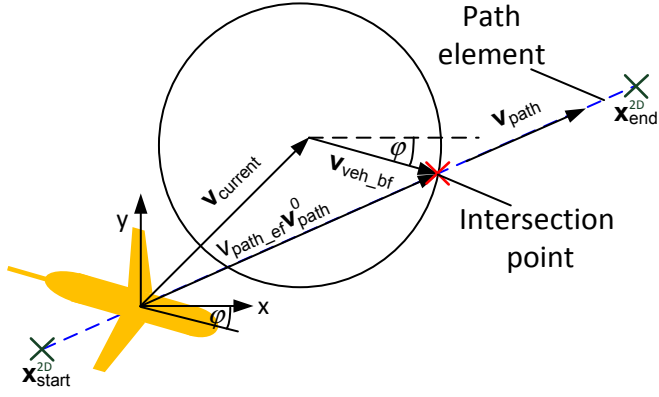

Fig. 2 Definition of the velocities

TABLE IV
ALGORITHM TO DETECT THE HEADING BY CALCULATION

$\varphi$ = CALC-HEADING($\mathbf{x}_{start\_path}$, $\mathbf{x}_{end\_path}$, $t_{start}$, $t_{end}$, $z_{climb-to}$, $z_{dive-to}$)
defined parameters: $v_{veh\_bf}$
$\mathbf{x}_{start\_climb\_to} = \mathbf{x}_{start\_path}$
$\mathbf{x}^z_{start\_climb\_to} = z_{climb-to}$
$\mathbf{x}_{start\_dive\_to} = \mathbf{x}_{start\_path}$
$\mathbf{x}^z_{start\_dive\_to} = z_{dive-to}$
$\mathbf{x}_{end\_climb\_to} = \mathbf{x}_{end\_path}$
$\mathbf{x}^z_{end\_climb\_to} = z_{climb-to}$
$\mathbf{x}_{end\_dive\_to} = \mathbf{x}_{end\_path}$
$\mathbf{x}^z_{end\_dive\_to} = z_{dive-to}$
$\mathbf{v}_{current\_start\_climb\_to}$ = GET-CURRENT($\mathbf{x}_{start\_climb\_to}$, $t_{start}$)
$\mathbf{v}_{current\_start\_dive\_to}$ = GET-CURRENT($\mathbf{x}_{start\_dive\_to}$, $t_{start}$)
$\mathbf{v}_{current\_end\_climb\_to}$ = GET-CURRENT($\mathbf{x}_{end\_climb\_to}$, $t_{end}$)
$\mathbf{v}_{current\_end\_dive\_to}$ = GET-CURRENT($\mathbf{x}_{end\_dive\_to}$, $t_{end}$)

$$\mathbf{v}_{current} = 0.25(\mathbf{v}_{current\_start\_climb\_to} + \mathbf{v}_{current\_start\_dive\_to} + \mathbf{v}_{current\_end\_climb\_to} + \mathbf{v}_{current\_end\_dive\_to}) \quad (6)$$

$$\mathbf{v}^0_{path} = \frac{\mathbf{x}^{2D}_{end} - \mathbf{x}^{2D}_{start}}{\|\mathbf{x}^{2D}_{end} - \mathbf{x}^{2D}_{start}\|} \quad (7)$$

$$disc = \left(\mathbf{v}^{0\,T}_{path} \cdot \mathbf{v}_{current}\right)^2 + v^2_{veh\_bf} - \mathbf{v}^T_{current} \cdot \mathbf{v}_{current} \quad (8)$$

if $disc < 0$ $disc = 0$

$$v_{path\_ef} = \mathbf{v}^{0\,T}_{path} \cdot \mathbf{v}_{current} + \sqrt{disc} \quad (9)$$

$$\mathbf{v}_{veh\_bf} = v_{path\_ef}\mathbf{v}^0_{path} - \mathbf{v}_{current} \quad (10)$$

$$\varphi = \mathrm{atan2}(\mathbf{v}^y_{veh\_bf}, \mathbf{v}^x_{veh\_bf}) \quad (11)$$

return $\varphi$

### B. Heading Simulation method

The heading simulation method is based on the simulation to determine the travel time $t^i_{path}$ for the path elements which are used in the path planning algorithms. This simulation includes a simplified dive profile of a glider to determine its behavior in several depths (The exact simulation of the saw-tooth glider dive profile is computationally time-intensive and impractical in path planning with one hundred thousand to one million discovered path elements). To include the position and depth-varying ocean current information in the simulation, the path element is divided into several path segments (see Fig. 3).

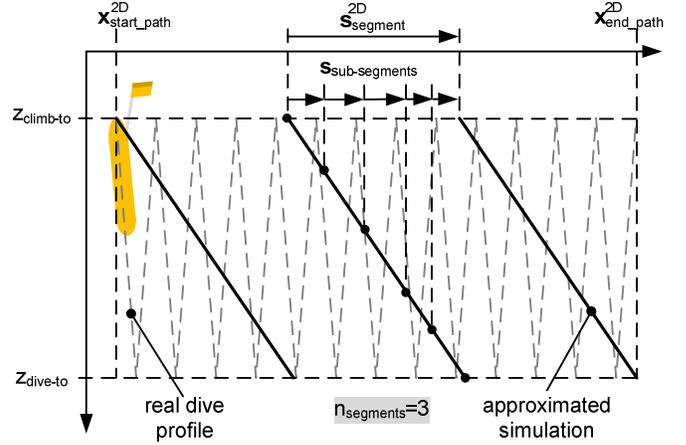

Fig. 3. Simplified dive profile along a path element

In each segment, the travel time will be calculated in function TRAVELTIME starting from the "climbto" depth $z_{climb-to}$ until the "diveto" depth $z_{dive-to}$. Detailed information about this function is described in Section III.B in [11]. The function TRAVELTIME is based on a step size control for efficient calculation of numerical solutions of differential equations. The step size $h$ here is not the time as used in numerical solvers but is a sub-segment of the segment. This function will be extended by the heading calculation of (10) and (11) to simulate the necessary heading commands for the glider to follow the path element.

Therefore, the segment will be shared in the function TRAVELTIME within many sub-segments for which equation (8)-(11) are solved. The current $\mathbf{v}_{current}$ is the arithmetic mean of the two velocities at the beginning and the end of the individual shared sub-segments. The calculated heading $\varphi$ for every sub-segment will be logged. At the end of the simulation a signal sequence $\boldsymbol{\varphi}$ of the heading is obtained. TABLE V includes the imported steps to detect the heading sequence.

TABLE V
PSEUDO-CODE OF THE ALGORITHM TO DETECT THE HEADING SEQUENCE

$\boldsymbol{\varphi}$=HEADING-SEQUENCE($\mathbf{x}_{start\_path}$, $\mathbf{x}_{end\_path}$, $t_{start}$, $t_{end}$, $z_{climb-to}$, $z_{dive-to}$)
defined parameters: $n_{segments}$
$\mathbf{s}^{2D}_{segment} = \left(\mathbf{x}^{2D}_{end\_path} - \mathbf{x}^{2D}_{start\_path}\right)/n_{segments}$
$t_{start\_segment} = t_{start}$
$\mathbf{x}^{2D}_{start\_segment} = \mathbf{x}^{2D}_{start\_path}$
$\mathbf{x}^z_{start\_segment} = z_{climb-to}$
$\mathbf{x}^z_{end\_segment} = z_{dive-to}$
$\boldsymbol{\varphi} = []$;
**for** ($i = 1$) **to** ($i = n_{segments}$)
 $\mathbf{x}^{2D}_{end\_segment} = \mathbf{x}^{2D}_{start\_segment} + \mathbf{s}^{2D}_{segment}$
 $\boldsymbol{\varphi}_i, t_{travel}$ = TRAVELTIME($\mathbf{x}_{start\_segment}$, $\mathbf{x}_{end\_segment}$, $t_{start\_segment}$)
 $t_{start\_segment} = t_{start\_segment} + t_{travel}$
 $\mathbf{x}^{2D}_{start\_segement} = \mathbf{x}^{2D}_{end\_segement}$
 $\boldsymbol{\varphi} = [\boldsymbol{\varphi}\ \boldsymbol{\varphi}_i]$
**return** $\boldsymbol{\varphi}$

From this sequence an average value of the heading $\varphi$ will be calculated and returned (see TABLE VI).

TABLE VI
PSEUDO-CODE OF THE ALGORITHM TO DETECT THE HEADING BY SIMULATION

| |
|---|
| $\varphi$=SIM-HEADING($\mathbf{x}_{start\_path}$, $\mathbf{x}_{end\_path}$, $t_{start}$, $t_{end}$, $z_{climb-to}$, $z_{dive-to}$) |
| $\boldsymbol{\varphi}$=HEADING-SEQUENCE($\mathbf{x}_{start\_path}$, $\mathbf{x}_{end\_path}$, $t_{start}$, $t_{end}$, $z_{climb-to}$, $z_{dive-to}$) |
| $\varphi$=mean($\boldsymbol{\varphi}$) |
| **return** $\varphi$ |

### C. Heading Optimization Method

This method detects the heading by solving an optimization problem. The optimization task is to find a commanded heading which leads the glider from the start point $\mathbf{x}_{start\_path}$ as closely as possible to the end point $\mathbf{x}_{end\_path}$. To simulate the glider behavior by means of a defined heading command the destination waypoint calculation in section IV.B will be used. The error $e$ is calculated by the Euclidean distance between the two dimensional end point $\mathbf{x}_{end}^{2D}$ and the simulated destination waypoint $\mathbf{x}_{dest}^{2D}$ in (12). The table below includes the steps of the error detection.

TABLE VII
PSEUDO-CODE OF THE ERROR FUNCTION

| |
|---|
| $e$ = ERROR-FUNCTION($\varphi$, $\mathbf{x}_{start\_path}$, $\mathbf{x}_{end\_path}$, $t_{start}$, $t_{end}$, $z_{climb-to}$, $z_{dive-to}$) |
| $t_{travel}=t_{end} - t_{start}$ |
| $\mathbf{x}_{dest}$=CALC-DESTINATION($\mathbf{x}_{start\_path}$, $t_{start\_start}$, $\varphi$, $t_{travel}$, $z_{climb-to}$, $z_{dive-to}$) |
| $e = \left\| \mathbf{x}_{end}^{2D} - \mathbf{x}_{dest}^{2D} \right\|$       (12) |
| **return** $e$ |

To solve this one-dimensional optimization task, bracketing methods can be used. These algorithms work without derivatives and find the minimum through iterative decreasing of the interval until the desired tolerance $\varepsilon$ is achieved, wherein the minimum lies. In this application Golden section search [15], Fibonacci search [16] and Brent's method [17] were tested. Brent's method had the best performance and will be favored.

To find good starting conditions for the optimization the minimal and the maximal value of the simulated heading sequence $\boldsymbol{\varphi}$ from the HEADING-SEQUENCE algorithm in TABLE V can be used as an initial interval. TABLE VIII includes the steps of this algorithm.

TABLE VIII
PSEUDO-CODE OF THE ALGORITHM TO DETECT THE HEADING BY OPTIMIZATION

| |
|---|
| $\varphi$ = OPT-HEADING($\mathbf{x}_{start\_path}$, $\mathbf{x}_{end\_path}$, $t_{start}$, $t_{end}$, $z_{climb-to}$, $z_{dive-to}$) |
| *defined parameters:* $\varepsilon$ |
| $\boldsymbol{\varphi}$=HEADING-SEQUENCE($\mathbf{x}_{start\_path}$, $\mathbf{x}_{end\_path}$, $t_{start}$, $t_{end}$, $z_{climb-to}$, $z_{dive-to}$) |
| $\varphi_{min}$ = min($\boldsymbol{\varphi}$) |
| $\varphi_{max}$ = max($\boldsymbol{\varphi}$) |
| $\varphi$ = BRACKETING-METHOD(ERROR-FUNCTION, $\varphi_{min}$, $\varphi_{max}$, $\varepsilon$) |
| **return** $\varphi$ |

## VI. RESULTS

### A. Comparision of the heading detection methods

This section presents the results of the individual heading detection methods to generate a DR waypoint list for glider missions. The Simple Simulator presented in section IV was used to simulate the position history of the glider. The focus of these tests is the comparison of the travelled glider paths using the DR waypoint list with the planned paths generated from a path planning algorithm in a time-varying ocean flow.

The function used to represent a time-varying ocean flow describes a meandering jet in the eastward direction, which is a simple mathematical model of the Gulf Stream [18] and [19]. This function was applied in [10, 11, 12] to test the TVE algorithm and its modifications, and in [20] to show the influence of uncertain information in path planning. The stream function is:

$$\phi(x,y) = 1 - \tanh\left( \frac{y - B(t)\cos(k(x-ct))}{\left(1 + k^2 B(t)^2 \sin^2(k(x-ct))\right)^{\frac{1}{2}}} \right) \quad (13)$$

which uses a dimensionless function of a time-dependent oscillation of the meander amplitude

$$B(t) = B_0 + \varepsilon \cos(\omega t + \theta) \quad (14)$$

and the parameter set $B_0 = 1.2$, $\varepsilon = 0.3$, $\omega = 0.4$, $\theta = \pi/2$, $k = 0.84$ and $c = 0.12$ to describe the velocity field:

$$u(x,y,t) = -\frac{\partial \phi}{\partial y} \quad v(x,y,t) = \frac{\partial \phi}{\partial x}. \quad (15)$$

The dimensionless value for the body-fixed vehicle velocity $v_{veh\_bf}$ is 0.5.

For the test cases, five different start positions were distributed in the whole area of operation. Fig. 10 shows the five paths found using the path planning algorithm, the generated DR waypoint list and the simulated glider track using the generated DR waypoint list. For the graph structure, the rectangular 3-sector grid structure with a grid size of 0.4 was used.

TABLE IX shows the summary of the test cases. The Position Error includes the Euclidean distance between the goal point and the destination position using the DR Waypoint list. The difference between the travel time calculated from the path planning and the simulated time is called Time Delay. Using the simplest method CAL, (Heading Calculation Method), a DR waypoint list cannot be generated for start point SP1. This is due to the complex current field and the insufficient adaptation of the commanded heading on the local changeable current conditions. The creation of an average current value for the whole path element like in (6) is not feasible for this test case. The generated DR waypoint lists using method SIM (Heading Simulation Method) and OPT (Heading Optimization Method) lead to successful missions. The simulated glider track and the planned path match very well. This shows the possibility of guiding a glider with single heading commands generated from a path planning waypoint list, although the path planning assumes that the vehicle holds the planned path using a permanent heading adaption. The position errors are the lowest when using the OPT method, which searches for an optimal heading to arrive at the destination waypoint as accurately as possible. The shorter travel times (negative time delays) can be explained by a very good adaptation of the glider tracks to the optimal solution.

TABLE IX
RESULTS OF THE DIFFERENT HEADING DETECTION METHODS

| Method | SP1 Position Error/ Time Delay | SP2 Position Error/ Time Delay | SP3 Position Error/ Time Delay | SP4 Position Error/ Time Delay | SP5 Position Error/ Time Delay |
|---|---|---|---|---|---|
| Travel Time | 17.7185 | 14.3213 | 11.6448 | 10.2084 | 3.8764 |
| CALC | -/- | 0.0084/ -0.0589 | 0.0027/ -0.0623 | 0.0003/ -0.0090 | 0.0003/ 0.0011 |
| SIM | 0.0103/ -0.2027 | 0.0001/ -0.0744 | 0.0027/ -0.0331 | 0.0006/ -0.0246 | 0.0002/ 0.0005 |
| OPT | 0.0075/ -0.2069 | 0.0001/ -0.0908 | 0.0009/ -0.0492 | 0.0001/ -0.0182 | 0.0002/ 0.0008 |

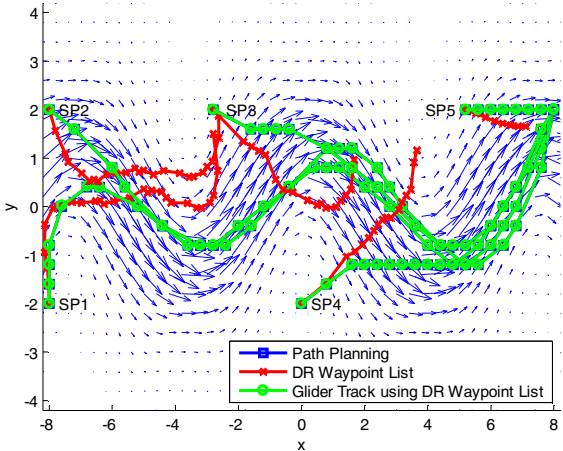

Fig. 4. Time optimal paths through a time-varying ocean field using path planning and a generated DR Waypoint List by using the OPT method

### B. Comparison the bracketing methods

The results achieved by using the investigated bracketing methods to detect an optimal heading command (see section V.C) will be presented in this section. The test cases correspond with the test scenario used in the previous section. The specific tolerance parameters for the methods were defined so that the methods delivered similar results. Brent's method shows the best results in all test cases with a performance-enhancement of over 30% compared to the other two methods and will be favored.

TABLE X
RESULTS OF THE DIFFERENT BRACKETING METHODS

| Method | SP1 No. of ErrorFcn Calls | SP2 No. of ErrorFcn Calls | SP3 No. of ErrorFcn Calls | SP4 No. of ErrorFcn Calls | SP5 No. of ErrorFcn Calls |
|---|---|---|---|---|---|
| No. of Path elements | 26 | 23 | 16 | 16 | 7 |
| Golden section search | 372 | 323 | 204 | 200 | 67 |
| Fibonacci search | 339 | 293 | 185 | 182 | 58 |
| Brent's method | 211 | 191 | 129 | 135 | 47 |

### C. Test the generated mission plans on the Glider Simulator

This section investigates the usability of the generated paths in practical glider missions. To reproduce a realistic glider behavior the Glider Simulator presented in section III will be used.

The function to simulate a realistic time–varying ocean flow is based on the dimensionless functions (13) to (15). As a result, the time scale corresponds to 3 days, and the space scales in x and y direction to 40 km. To include a depth dependence of ocean flow a time and depth variant term $u_{surface}(z,t)$

$$u_{surface}(z,t) = W(t) \max\left(\left(1 - \frac{1}{z_{max}}z\right), 0\right) \quad v(z,t) = 0 \quad (16)$$

$$W(t) = W_0 \cos(d\omega t) \quad (17)$$

with the parameters $z_{max}$ = 15 m, $W_0$ =0.5 and $d$ =2 will be added to the $u(x,y,t)$ component of (15):

$$u(x,y,z,t) = u(x,y,t) + u_{surface}(z,t)$$
$$v(x,y,z,t) = v(x,y,t) \quad . \quad (18)$$

This additional term shall describe the influence of the wind on the surface current. This influence decreases linearly up to a depth $z_{max}$ as well as being time variant with an angular frequency $d\omega$ (see (17)). As a result, $u_{surface}$ emulates the behaviour of a head- (negative values for $W(t)$) and tailwind (positive values for $W(t)$) in x-direction.

The following table includes the important parameters to generate a time optimal path using an A*TVE algorithm [12].

TABLE XI
PATH PLANNING PARAMETERS

| Settings | |
|---|---|
| Path Planning Method | A*TVE |
| Grid Size in x and y Direction | 5000 m |
| Graph Structure | Rectangular 3-sector |
| Operation Area x Dimension | [-320 – 320] km |
| Operation Area y Dimension | [-160 – 160] km |
| Tactical Data | |
| Vehicle speed DR calculation | 0.342 m/s |
| Vehicle Speed Glider Simulation | 0.382 m/s |
| Climb to Depth | 2.52 m |
| Dive to Depth | 102.48 m |
| Start Time | 2.378 d |
| Start Position | [0, 0] km |
| Goal Position | [125, -100] km |
| Benchmark | |
| No. of Vertices | 8385 |
| No. of Edges | 257936 |
| No. of Cost Function Calls | 7290 |
| No. of Current Model Calls | 545069 |
| Computing Time (Intel® Core™ i7-4900MQ) | 0.213 s |
| Results | |
| Length Unsmoothed Path | 168.6 km |
| Length Smoothed Path | 168.255 km |
| No of Waypoints Unsmoothed Path | 17 |
| No of Waypoints Smoothed Path | 6 |
| Travel Time Unsmoothed Path | 2.459 d |
| Travel Time Smoothed Path | 2.456 d |

The extraction of the tactical data for the glider occurs by analysis of a simple test mission which consists of a start and a goal position. The ocean current influence was set in the simulator with a speed of 0 m/s. The commanded "diveto" and "climbto" depth in the glider mission plan was [5 100] m. The analyses of the logged dive profiles have produced an average range from [2.52 102.48] m. This is the result of an overshoot of 2.5 m in the depth control. The distances and the time delays between consecutive "diveto" peaks of the saw-tooth glider dive profile were analyzed to detect the vehicle speed. It was discovered that the dead reckoning algorithm and the Glider Simulator produce different vehicle speeds. To take account of this fact, two velocities will be used in the path planning. The vehicle speed in the DR algorithm will be used to create the DR waypoint list. This leads to a synchronous switching to the next heading command between the planned and the simulated time. The vehicle speed of the Simulator will be used in the path planning algorithm to calculate the cost functions for the edges and for the heading detection methods presented in section V.

Fig. 5 and Fig. 6 show the simulated glider tracks using DR waypoint lists generated from an unsmoothed and smoothed path using the OPT method presented in section V.C.

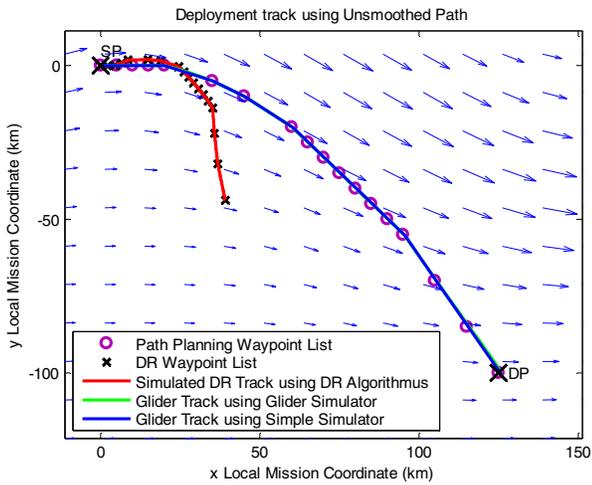

Fig. 5. The development track of an unsmoothed path from path planning

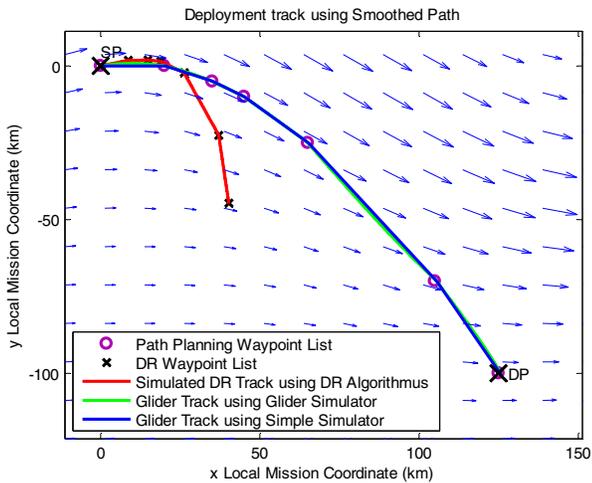

Fig. 6. The development track of a smoothed path from path planning

The glider tracks are very well in accordance with the planned paths. Another point of interest is the accuracy of the Simple Simulator (blue line) in comparison to the Glider Simulator (green line). This is a necessary requirement to generate the correct heading commands for the DR waypoint list.

TABLE XII includes the results using the heading detection methods to generate the DR waypoint lists. The CALC method has the largest position errors and the longest time delays. The reason for this is the insufficient approximation of the glider behavior using only one average ocean current vector for the whole path element, where the ocean current is not constant in time and space. The original paths found are unsmoothed and characterized by many path elements with change of directions. To decrease the number of path elements without increasing the travel time, a path smoothing algorithm will be used [11]. The fewer waypoints in the smoothed path also lead to a larger position error due to the insufficient reproduction of the time optimal path. The reached position error is less than 1 km and the time delay to the path planning solution is approximately 10 min using the Glider Simulator. This is an excellent result considering that the glider dives the whole mission without GPS updates. It is clear that these results are attributable to the fact that an identical ocean current model is used in the Glider Simulator and in the path planning. In real missions where the path planning uses the ocean current information provided from a forecast system there exist additional uncertainties. The more accurate results using the Simple Simulator for mission simulations, compared to the Glider Simulator, are attributed to the fact that the same vehicle behavior algorithms and tactical data are used in the path planning, heading detection algorithm, and in the Simple Simulator. Using the Glider Simulator estimated parameters for the vehicle speed, the "climbto" and "diveto" depths are used which can be interpreted as uncertainties.

The Glider Simulator allows realistic simulations of glider missions for complex test scenarios and a fast availability of the test results. The complexities of the planned paths, as well as the accurate work of the Glider Simulator are demonstrated in Fig. 7 and Fig. 8. These plots include the logged ocean current information during the mission. The simple simulation algorithms in the path planning reproduce these complex profiles very well to calculate the travel times for the path elements. The exact reproduction of the saw-tooth dive profile in the Glider Simulator can be well seen at the peaks.

TABLE XII
RESULTS OF THE DIFFERENT HEADING DETECTION METHODS

| Path | Method | Simple Simulator | | Glider Simulator | |
|---|---|---|---|---|---|
| | | Position Error in m | Time Delay in min | Position Error in m | Time Delay in min |
| Unsmoothed | CALC | 1807.15 | 20.32 | 977.34 | 20.18 |
| | SIM | 266.89 | 5.21 | 611.34 | 9.56 |
| | OPT | 26.14 | 4.46 | 844.51 | 8.16 |
| Smoothed | CALC | 3303.17 | 53.73 | 2691.66 | 52.46 |
| | SIM | 649.37 | 5.57 | 122.97 | 8.58 |
| | OPT | 126.11 | 7.55 | 676.07 | 12.19 |

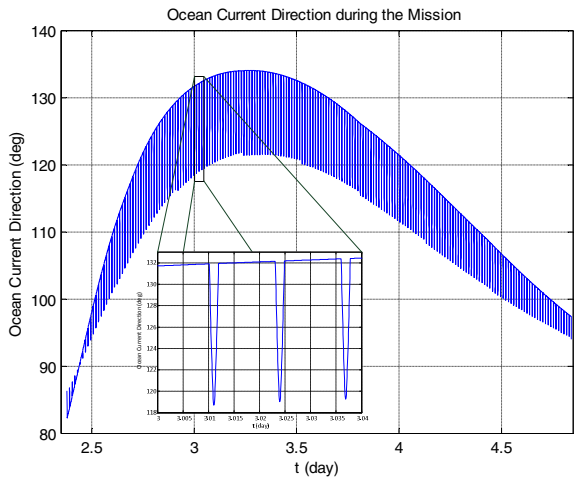

Fig. 7. Ocean current direction during the mission

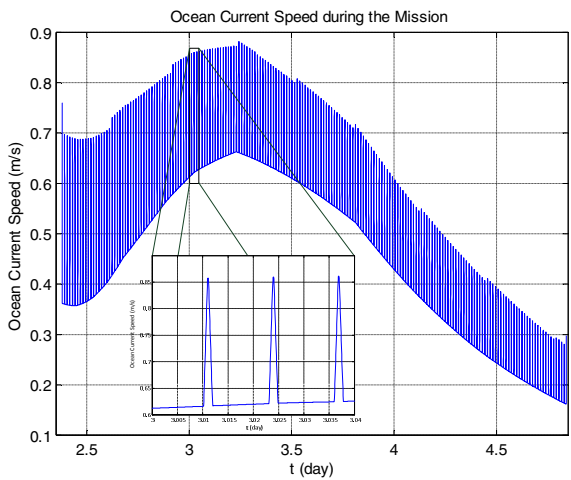

Fig. 8. Ocean current speed during the mission

## VII. CONCLUSIONS AND FUTURE WORK

In this paper, mechanisms for performing path planning using the glider software are presented. A significant challenge of this work is the inaccurate detection of the vehicle's position during a mission. The onboard dead reckoning algorithm delivers insufficient results in the case of time-varying and depth-dependent ocean flows when the current correction system is enabled. Hence, a DR waypoint list based on heading commands is generated using the planned time-optimal path that considers all information about ocean conditions during the mission. For our tests, we used a faster-than-real-time Glider Simulator which is based on the standard glider control software from Teledyne Webb Research. This tool was essential for the development and testing of the algorithms. The Glider simulator allows for realistic reproduction of missions with inclusion of uncertainties, such as varying vehicle speeds. This is useful for our future research, as path planning algorithms should include uncertain information, like the usage of multiple ocean forecast models, in order to feed the path planning system source information that is as reliable as possible [21].